# CREATION OF A CHATBOT BASED ON NATURAL LANGUAGE PROCESSING FOR WHATSAPP

Valderrama Jonatan[1] and Aguilar- Alonso Igor [1,2]

[1]Faculty of Systems Engineering and Informatics, National University of San Marcos Lima, Peru
[2]Professional School of Systems Engineering, National Technological University of South Lima

## ABSTRACT

*In the era of digital transformation, customer service is of paramount importance to the success of organizations, and to meet the growing demand for immediate responses and personalized assistance 24 hours a day, chatbots have become a promising tool to solve these problems. Currently, there are many companies that need to provide these solutions to their customers, which motivates us to study this problem and offer a suitable solution. The objective of this study is to develop a chatbot based on natural language processing to improve customer satisfaction and improve the quality of service provided by the company through WhatsApp. The solution focuses on creating a chatbot that efficiently and effectively handles user queries. A literature review related to existing chatbots has been conducted, analyzing methodological approaches, artificial intelligence techniques and quality attributes used in the implementation of chatbots. The results found highlight that chatbots based on natural language processing enable fast and accurate responses, which improves the efficiency of customer service, as chatbots contribute to customer satisfaction by providing accurate answers and quick solutions to their queries at any time. Some authors point out that artificial intelligence techniques, such as machine learning, improve the learning and adaptability of chatbots as user interactions occur, so a good choice of appropriate natural language understanding technologies is essential for optimal chatbot performance. The results of this study will provide a solid foundation for the design and development of effective chatbots for customer service, ensuring a satisfactory user experience and thus meeting the needs of the organization.*

## 1. INTRODUCTION

The customer service area plays a critical role in the success of any organization. With the constant growth of e-commerce and the need for immediate feedback, it is important to provide users with a satisfying and efficient experience. In this context, chatbots seem to be a promising tool to provide automated and personalized support.This study focuses on developing a chatbot based on natural language processing for WhatsApp, with the purpose of improving customer satisfaction and service quality. The existing literature in the field of chatbots was reviewed in detail, analysing methodological approaches, artificial intelligence techniques and quality attributes used in the implementation of these systems. The literature has highlighted that chatbots based on natural language processing allow fast and accurate responses, which translates into a significant improvement in customer service efficiency [1]. Furthermore, chatbots have been observed to contribute to customer satisfaction by providing accurate responses and quick solutions to their queries [1]. Therefore, it is of vital importance to design a chatbot with a friendly interaction and that mimics human interactions, to offer a satisfactory user experience [4].

Another consideration is choosing the right technology to understand natural language. Essential for good chatbot performance[2]. Previous research has highlighted the importance of using artificial intelligence techniques such as machine learning to improve chatbot learning and its adaptability when interacting with users.

Developing an effective customer service chatbot requires not only advanced tools and technology implementation, but also a deep understanding of user needs and expectations.
Through a comprehensive review of the literature, methodological approaches, artificial intelligence techniques and quality attributes relevant to the successful implementation of the chatbot will be identified.

## 2. THEORETICAL FRAMEWORK

### 2.1. Chatbot

Chatbot is an application that simulates human conversation in a chat interface. It uses advanced artificial intelligence and natural language processing techniques to understand and provide automated responses to user queries. Chatbots find applications in a variety of scenarios, including customer service, help desk, sales, and marketing. They offer a convenient and efficient way to interact with users, simulating a human-like conversation while taking advantage of intelligent algorithms and language processing capabilities.

The historical evolution of chatbots is essential to understand their development and current applications. Several studies have investigated this evolution and offer an overview of significant milestones and advances in this field [1]. From early rule-based systems to sophisticated AI-based chatbots, there has been a remarkable growth in the power and versatility of chatbots [2].

### 2.2. Types of Chatbots

Chatbots are classified into different types based on their features and functionality. Below are the main types of chatbots:

1. **Rule-based chatbots**. Work by applying predefined instructions and providing predetermined responses based on specific input patterns. These chatbots are efficient in situations where the queries are clear, and a limited set of responses are available [3]. However, its limitation lies in the difficulty of dealing with ambiguous or complex queries.
2. **Chatbots based on Artificial Intelligence**. Leveraging methods like natural language processing and machine learning, chatbots have the capacity to comprehend and generate responses in human language [4].These AI-driven chatbots possess the ability to learn and enhance their performance over time through interactions with users. This enables them to be versatile in various scenarios, delivering responses that are not only more precise but also contextually relevant.
3. **Voice chatbots**. Are designed to interact using voice commands. These chatbots use speech recognition and synthesis technologies to understand and generate spoken responses [6]. Voice chatbots are especially useful in situations where the use of hands or vision is limited, such as in automotive applications or smart home devices.
4. **Hybrid chatbots**. Combine features of rule-based and AI-based chatbots. These chatbots use predefined rules for common cases and resort to artificial intelligence techniques in more complex situations [5]. This blend of capabilities enables a higher degree of adaptability and receptiveness in customer service.

Every variety of chatbot comes with its own set of advantages and constraints, and the selection hinges on the requirements and goals of the company.

### 2.3. Natural Language Processing

Natural language processing (NLP) is a fundamental field in the development of intelligent chatbots. Various approaches and models related to NLP have been proposed, such as transformer-based models, which have shown excellent results in understanding and generating natural language [7] .

These models use tokenization, attention, and decoding techniques to improve the chatbots' ability to understand queries and generate appropriate responses.

On the other hand, machine learning plays an important role in the development of intelligent chatbots for fluid and contextual conversations [8]. Machine learning techniques are used for consistent response generation, user intent detection, and dialog personalization. These approaches allow chatbots to adapt to the preferences and needs of users, thus improving the quality of interaction.

### 2.4. WhatsApp

Is an instant messaging application that enables users to send text messages, initiate voice, and video calls, share various files and multimedia content, as well as engage in group conversations. It was developed as a communication platform for smartphones and has become one of the most popular messaging apps in the world.

### 2.5. Integration of Chatbots in Customer Service

The effective integration of chatbots in customer service is an important aspect to consider. Strategies and best practices are explored to implement chatbots in different customer service channels, such as online chat, social networks, and mobile applications [8]. In addition, the personalization of responses is key to providing a more satisfactory experience, adapting interactions to the individual preferences and needs of each customer. Finally, it matters

## 3. METHODOLOGY

For the development of this research, we used [26] guide, which establishes three important parts.

1. Planning. This phase is important to consider the requirements for carrying out the literature review, considering the information search sources, the research questions and the search criteria.
2. Conduct of the review. In this phase, the methodological selection of the information from the main studies is carried out according to the inclusion and exclusion criteria.
3. Results of the review. In this phase, the statistical results of the studies selected for the literature review in each of the information sources are presented in summary. These results will serve for our research proposal.

To develop this research, a literature review of scientific articles no older than 5 years was carried out, extracted from important scientific databases such as Science Direct, Springer Link, Emerald Insight, IOP science and Taylor & Francis Online.

To learn better about the types of interaction with users, the AI techniques and algorithms used, the attributes, the technologies used in development, the mechanisms for training chatbot data, we review articles from different authors of research related to chatbots. In the area of education, it was necessary to ask the research questions indicated below:

Q1 What are the types of user interaction with the chatbot?
Q2 What artificial intelligence techniques and algorithms will be employed in the chatbots development?
Q3 What will be the quality attributes of the chatbot?
Q4 Which technologies will be utilised for the chatbots development?
Q5 What mechanism will be adopted to train the chatbot data?

According to the articles identified in the literature review process according to the established search string, the articles were filtered according to the inclusion and exclusion criteria of Table 1, resulting in many of them excluded for not meeting the established criteria. Other articles were excluded because they did not contribute significantly to our research.

Table 1. Inclusion and exclusion criteria

| Inclusion criteria | Exclusion criteria |
| --- | --- |
| Articles published from 2018 to 2023 | Articles that are not in the 2018, 2023 range |
| Articles in English or Spanish | Articles written in languages other than English or Spanish |
| Articles related to chatbots for customer service | Other issues unrelated to customer service |

The search string was extensively designed using key terms related to chatbots, user interaction, artificial intelligence techniques, quality attributes and the technologies used in its development. Searches were performed on different combinations of implement chatbots in different customer service channels, such as online chat, social networks, and mobile applications [8]. Furthermore, the benefits and Keywords.

Table 2. Search string applied in the databases.

| Databases | Search Strings |
| --- | --- |
| Science Direct | natural language processing, chatbot implementation, intelligent conversational agents, intelligent conversational agents |
| SpringerLink | natural language processing, chatbot development, conversational AI |
| Emerald Insight | natural language processing, chatbot applications, conversational agents |
| IOPscience | natural language processing, chatbot algorithms, intelligent dialogue systems |
| Taylor & Francis Online | natural language processing, chatbot evaluation, conversational interfaces |

After applying the inclusion and exclusion criteria, a total of forty-eight potential studies were obtained that could provide relevant information to answer the research questions posed in the study. These studies were carefully examined, reviewing their titles, abstracts, and the full content of each article.

From this review process, thirty-five relevant studies were identified that directly addressed the research questions and provided valuable information on the types of user interaction with chatbots, AI techniques and algorithms used, chatbot quality attributes, the technologies used in its development and the data training mechanisms.

Finally, twenty-five selected studies were considered based on their relevance to the research topic, the solidity of their methodology and their valuable contributions to the field of study.

Table 3. Articles found and selected by source consulted.

| Database | Potential studies | Relevant studies | Selected studies | % |
|---|---|---|---|---|
| Science Direct | 15 | 15 | 15 | 56% |
| SpringerLink | 12 | 10 | 5 | 19% |
| Emerald Insight | 6 | 3 | 2 | 7% |
| IOPscience | 6 | 3 | 2 | 7% |
| Taylor & Francis Online | 9 | 6 | 3 | 11% |
| TOTAL | 48 | 35 | 25 | 100% |

## 4. DESCRIPTION OF THE RESULTS

### 4.1. Types of user Interaction with Chatbot

To answer this question, we consulted several articles that examine the types of user interaction with chatbots. Among them are the following:

The user communicates with the chatbot by sending text messages and receiving responses in text form. This form of interaction is widely used in chatbots, as it is simple and accessible to most users, according to the study by [24].

The user communicates with the chatbot using voice commands and receives spoken responses. This form of interaction has become more popular with the advancement of voice recognition technology, according to the study by [16].

The chatbot allows the user to interact through text entry and voice commands, providing the flexibility to choose the user's preferred method, as indicated in the study by [20].
The chatbot presents predefined options in the form of buttons or drop-down menus, allowing the user to select an option and receive responses according to their choice, as mentioned in the study by [21].

The user asks the chatbot specific questions and receives direct answers related to the query. This type of interaction focuses on getting clear and direct answers to the user questions, as mentioned in the study by [2].

These reviewed articles provide a solid foundation for understanding the different forms of user interaction with chatbots, allowing us to recognise and understand the basic characteristics of the interaction, which can be very valuable when creating our chatbot.

## 4.2. Artificial Intelligence Techniques and Algorithms Employed in the Chatbots Development

Techniques and algorithms are vital for creating an efficient chatbot. Several pertinent techniques and algorithms are identified from the analysis of related literature.
[1] suggest using techniques such like intent matching, machine learning, and natural language processing.

Additionally, the study conducted by [2] highlights the use of metamodels and natural language processing.

These results demonstrate the importance of using techniques and algorithms such as machine learning, intention classification, and sentiment analysis to enhance chatbot efficiency and precision.

## 4.3. Chatbot Quality Attributes

To identify quality attributes of the chatbot, we analysed numerous studies on chatbot usability and user experience.

The use of chatbots to support educational systems was identified, as mentioned in the study by [6], highlights the importance of usability, ease of use and user satisfaction. Furthermore, article [14] mentions the impact of "humanizing" chatbots to improve user satisfaction.

These articles offer a strong basis for examining quality attributes such as effectiveness, efficiency, usability, user satisfaction, and responsiveness when developing chatbots.

## 4.4. Technologies for the Development of Chatbots

Selecting suitable technologies is a crucial factor in the successful development of a chatbot. By analysing the relevant articles, the key technologies used in chatbot development are identified.
Article [1], discusses the utilization of neural networks, natural language processing, machine learning, and chatbots. Moreover, the study by Abdellatif et al. [2] highlights the use of natural language understanding platforms for the development of chatbots.

These findings highlight the relevance of technologies like neural networks, natural language processing, and machine learning in the creation of efficient and effective chatbots.

## 4.5. The Mechanism for Training the Chatbot Data

The process of training chatbot data is fundamental to achieving high accuracy and performance of a chatbot. By reviewing the relevant articles, different mechanisms used to train data in chatbot development are identified.

In particular, [24] provide a comparison of natural language understanding platforms for chatbots in software engineering. They analysed the performance of different platforms using supervised learning techniques and evaluated their ability to understand user queries in the context of software engineering. Additionally, [16] explored the effects of AI-based chatbots on user compliance in customer service. They employed a supervised learning approach to train the chatbots and evaluated their impact on user behaviour and satisfaction, particularly concerning adherence to the instructions provided by the chatbot.

[17] conducts a comparative analysis of the performance of a multimodal chatbot implementation that utilises on news classification via categories.

## 5. ANALYSIS OF THE RESULTS

### 5.1. Types of User Interaction with the Chatbot

We can examine in greater detail the percentages associated with each type of interaction, as shown in Figure 1.

Interaction I1, which is based on the use of text, represents 37.5% of the total interactions studied. This remarkable figure underscores the prevalence of textual communication in the context of chatbots. The authors related to this interaction are: [24], [14] and [20].

Interaction I2, which involves the use of voice, constitutes 12.5% of the interactions. This finding highlights the growing adoption of speech recognition technology and its integration into chatbot systems. The author related to this interaction is [16].

Interaction I3, which combines the use of text and voice, also occupies 12.5% of the analysed interactions. This convergence of communication modalities demonstrates the importance of offering multiple and flexible options to users. The author related to this interaction is [20].

The I4 Interaction, based on the use of buttons, also represents 12.5% of the interactions. This result suggests the relevance of an intuitive and simplified user interface. The author related to this interaction is [13].

Interaction I5, which is based on a question-answer model, also shows a significant presence, representing 25% of the interactions studied. This emphasizes the importance of chatbots' ability to provide accurate and relevant responses to user queries. The authors related to this interaction are [22] and [13].

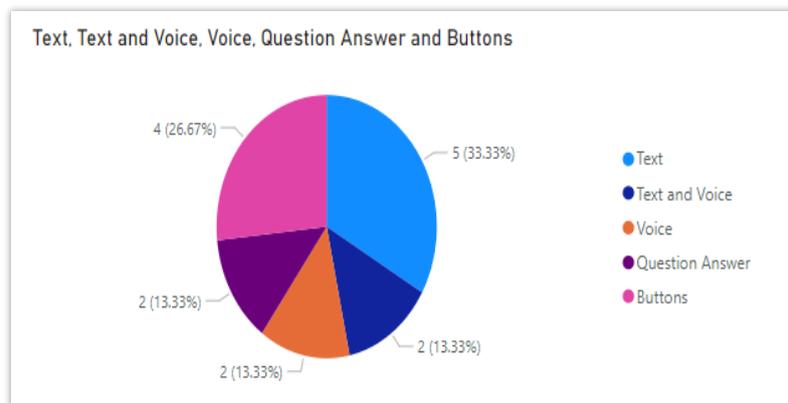

Figure 1. Types of interaction

### 5.2. Artificial Intelligence Techniques and Algorithms Used

Figure 2 shows the results found of the AI techniques and algorithms that are used to develop chatbots.

17

The Decision Trees technique as an algorithm represents 25% of the techniques used in chatbots. This indicates that the authors [24] and [14] have recognized the importance of using decision trees in the development of their chatbots. This finding suggests that these algorithms are effective in making decisions and generating appropriate responses for users.

The natural language processing (NLP) technique also represents 25% of the techniques used. This indicates that the authors [24] and [16] recognize the importance of understanding and processing natural language to achieve effective communication with users. This technology is crucial to understanding queries and generating consistent and meaningful responses.

The Support Vector Machines technique as an algorithm represents 12.5% of the techniques used. This implies that the author [21] has explored the use of these algorithms in their chatbots. Support Vector Machines are renowned for their capability to classify and analyse intricate data, a feature that could prove advantageous in the realm of chatbots.

The Recurrent Neural Networks (RNN) technique also represents 25% of the techniques used. This indicates that the authors [24] and [16] have recognized the utility of RNNs in the development of chatbots. Recurrent Neural Networks are recognized for their capacity to handle data streams, a characteristic that holds relevance in user conversation contexts.

The Markov Chain technique and Long Short-Term Memory (LSTM) as algorithms represent 12.5% of the techniques used.

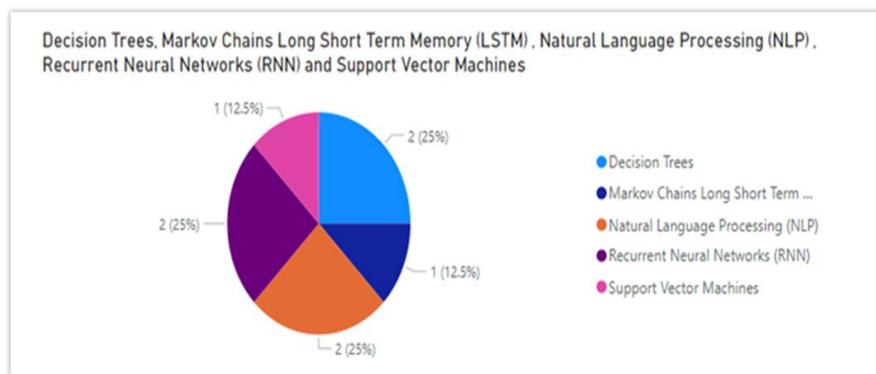

Figure 2. Artificial intelligence techniques and algorithms

## 5.3. Chatbot Quality Attributes

Several articles focused on the user experience and usability of chatbots have been analysed. Figure 3 provides a more detailed look at the specific attributes to consider when developing chatbots.

Naturalness: 42.86% of the authors have addressed naturalness in their research. This means that they have researched and considered the importance of chatbots being able to generate responses and conversations that are as natural and human as possible. This attribute seeks that users perceive the chatbot as an entity with which they can interact in a fluid and natural way.

Speed: Regarding speed, it is observed that an author [16] has considered this attribute in his research. Speed, in the context of chatbots, pertains to their capability to deliver prompt and efficient responses to user inquiries. A fast chatbot can enhance the user experience by furnishing information in a timely fashion.

Availability: One author[14] has addressed availability in research on it. Availability pertains to the chatbot's capacity to be accessible and ready for users at any given time. This means that the chatbot is available to answer questions and help at any time of the day.

Precision: Another author [13] has considered the precision in the investigation of it. Accuracy refers to the chatbot's ability to provide correct and exact answers. An accurate chatbot is capable of correctly understanding user queries and delivering accurate and relevant responses.

Learning: Lastly, it is important to mention that one of the authors [1] has explored the aspect of learning in their research.

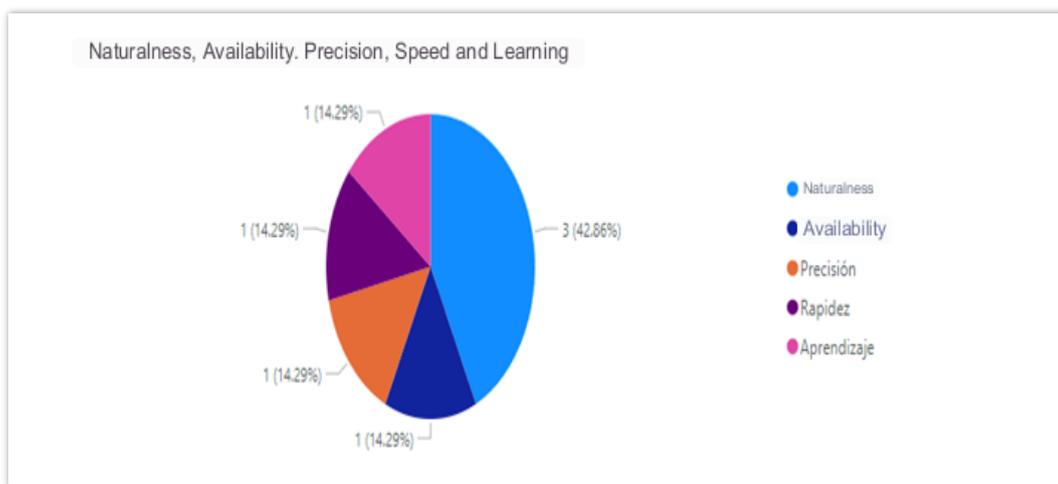

Figure 3. Quality attributes

## 5.4. Technologies for the Development of Chatbots

The following technologies are among the primary ones used in the development of chatbots, as shown in Figure 4.

Python: This technology, mentioned by the author [24] represents 10% of the total technologies used in chatbot development. Python is a widely used programming language in the realm of artificial intelligence and natural language processing, making it a popular choice for implementing chatbots.

Dialogflow: This technology, mentioned by the authors [24] and [20],represents 20% of the technologies used.

Dialogflow is a cloud-based chatbot development platform that provides advanced natural language processing and intent understanding capabilities.

Keras: This technology, mentioned by the author [24] represents another 20% of the technologies used. Keras is a high-level library for constructing and training neural networks, commonly employed in deep learning and natural language processing.

IBM Watson: This technology, mentioned in articles [24] and [19], represents a percentage of 20%. IBM Watson is an artificial intelligence and machine learning platform that offers a wide range of services for the development of chatbots and other AI-based applications.

Twilio: This technology, mentioned in articles [24], [17] and [1], also represents the highest percentage with 30% of the technologies used.

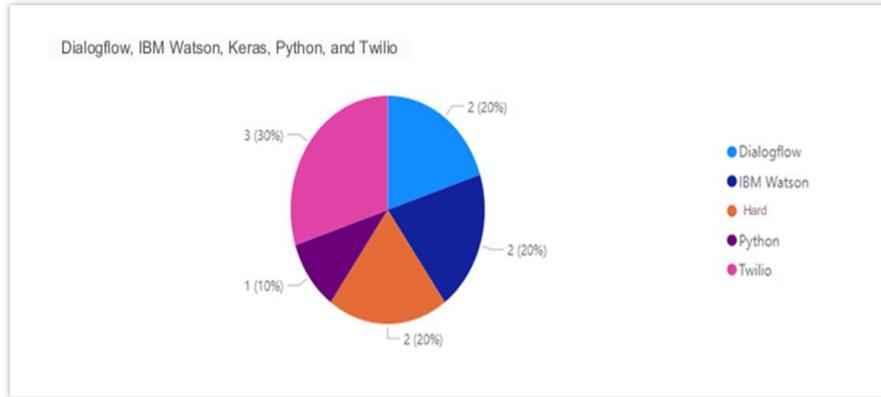

Figure 4. Development technologies

## 5.5. The Mechanism for Training the Chatbot Data

The results of the most appropriate mechanisms for data training in a chatbot can be seen in Figure 5.

Supervised Learning: 30% of the authors have addressed supervised learning in their research. [24] have used this approach in their work. Supervised learning entails training the chatbot using a labeled dataset, in which instances of anticipated input and output are provided. This allows the chatbot to learn to generate correct responses based on previous patterns and examples.

Reinforcement Learning: 10% of authors have explored reinforcement learning in their studies. [1] have investigated this approach in their work. Reinforcement learning involves the chatbot interacting with the environment and receiving feedback in the form of rewards or penalties. Through feedback, the chatbot learns to make decisions that maximize rewards over time.

Transfer of Learning: 10% of the authors have considered the transfer of learning in their research. [20] have mentioned this approach in their work. Transfer of learning involves drawing on a model trained on a task's prior knowledge and experience and applying it to a related but different task. This expedites the training process and enhances the chatbot's performance in the new task.

Generation of Synthetic Data: 10% of the authors have addressed the generation of synthetic data in their studies. In this case, a specific author has not been provided. Synthetic data generation involves creating artificially generated training data to increase the number and diversity of examples available to the chatbot. This can improve the chatbot's ability to generalize and handle a variety of situations.

Active Learning: 40% of the authors have investigated active learning in their work. [24] mention this approach in their article.

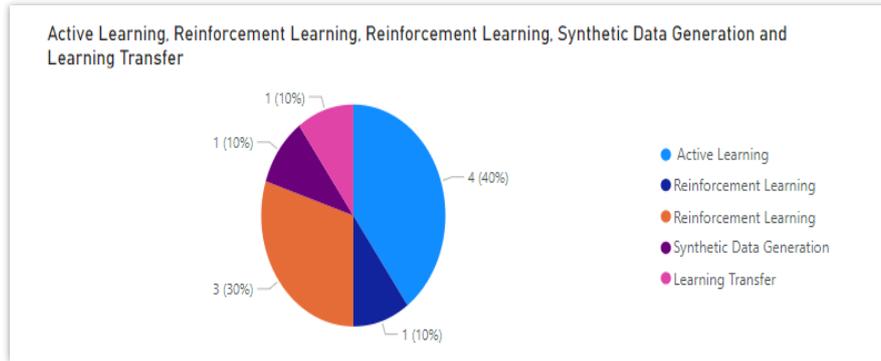

Figure 5: Data training mechanism

## 6. PROPOSED ARCHITECTURE FOR CHATBOT

The architecture of the system implemented in this project comprises three levels, as shown in Figure 6.

First level: involves the WhatsApp client application that the user will use to access the chatbot, Second level: We have the AI engine built with python and the Flask framework using natural language processing.

Third level: We have the Twilio-based REST API for communication between the front-end application and the AI engine.

The customer service module is responsible for interacting with users through the WhatsApp application. It uses an artificial intelligence engine that leverages natural language processing to successfully understand and respond to queries [1].

The backend module is divided into two submodules. The first submodule consists of a Twilio webhook application that enables communication between the front-end application and the AI engine, located in the second backend submodule [2]. This AI engine uses natural language processing algorithms and techniques implemented in a Python application. For the development of the system, backend frameworks such as Django or Flask are used, which facilitate the implementation of projects of this type [3].

The data set used contains information about the service company and is organized with common words, possible intentions, and the corresponding responses. Natural language processing techniques, such as a bag of words, are applied to count the frequency of words in the data set. In addition, pre-processing tasks such as tokenization and removal of symbols and special characters are performed. To build the natural language model, tools such as TensorFlow and Keras [4] are used.

The chatbot solution to satisfy customer inquiries requires an internet connection and an Android mobile device with the WhatsApp app installed. Users initiate the query flow by typing the number provided by Twilio [5].

The AI engine employs a fully connected neural network architecture, known as dense layers, for intent classification. Dense layers are used with the ReLU (Rectified Linear Unit) activation function to learn patterns and nonlinear representations in the input data. The model is trained using the SoftMax activation function in the output layer to assign probabilities to each intention class [6]. The chatbot architecture consists of client, AI, and REST API layers. The interaction is done through the WhatsApp application and the AI engine processes the queries using natural language processing techniques. The natural language model is constructed using dense layers within a neural network, employing the SoftMax activation function for intent classification.

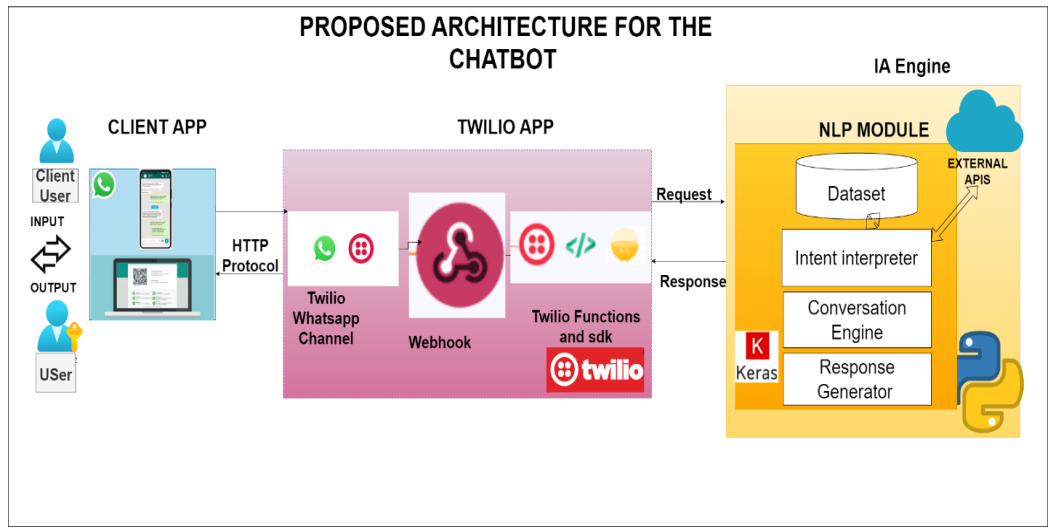

Figure 6: Proposed architecture for the chatbot

## 7. CONCLUSIONS

In our research, we reviewed several studies related to chatbots that provide customer service, focusing on their efficiency, customer satisfaction, and the quality of the service they provide.
The results revealed that chatbots based on natural language processing improve customer service efficiency by providing fast and accurate responses.

Chatbots were also found to contribute to customer satisfaction by providing quick and accurate solutions to their problems.

The studies found have provided information on the importance of investing in technologies and tools that support natural language processing and artificial intelligence, as well as user-centred design to improve user experience and satisfaction.

According to the literature review, a chatbot architecture has been proposed through WhatsApp that is friendly and personalized to achieve a positive interaction with the user.